\documentclass[letterpaper, 10 pt, conference]{ieeeconf}
\IEEEoverridecommandlockouts
\usepackage[space, compress, sort]{cite}
\usepackage{amsmath}
\usepackage{amsfonts}
\usepackage{diagbox}

\usepackage{amsthm}
\usepackage[ruled,vlined]{algorithm2e}
\usepackage{mathtools}
\usepackage{tabularx}
\usepackage{graphicx}
\usepackage{enumerate}
\usepackage{float}
\usepackage{url}
\usepackage{verbatim}
\usepackage{booktabs} 
\usepackage{multirow}
\usepackage[linkcolor=black,citecolor=black,urlcolor=black,colorlinks=true]{hyperref}
\usepackage{bm}
\DeclareMathOperator*{\argmin}{arg\,min}

\newcommand{\abs}[1]{\vert{#1}\vert}

\newcommand\add[1]{\textcolor{black}{#1}}
\usepackage{algpseudocode}

\newtheorem{definition}{Definition}

\newcommand{\tabincell}[2]{\begin{tabular}{@{}#1@{}}#2\end{tabular}}
\title{ \LARGE \bf Trajectory Optimization with Global Yaw Parameterization for Field-of-View Constrained Autonomous Flight
}
	\author{Yuwei Wu, Yuezhan Tao, Igor Spasojevic, Vijay Kumar
	\thanks{The authors are with the GRASP Laboratory, University of Pennsylvania, Philadelphia, PA, 19104 USA {\tt\small\{yuweiwu, yztao, igorspas, kumar\}@seas.upenn.edu}.}
   \thanks{We gratefully acknowledge the support of The Institute for Learning-Enabled Optimization at Scale (TILOS) funded by the National Science Foundation (NSF) under NSF Grant CCR-2112665, IoT4Ag ERC funded through NSF Grant EEC-1941529, ONR grant N00014-20-1-2822, and ONR grant N00014-20-S-B001. We also thank Jake Welde for providing helpful insights.}}
\begin{document}
\maketitle

\begin{abstract}

Trajectory generation for quadrotors with limited field-of-view sensors has numerous applications such as aerial exploration, coverage, inspection, videography, and target tracking. 
Most previous works simplify the task of optimizing yaw trajectories by either aligning the heading of the robot with its velocity, or potentially restricting the feasible space of candidate trajectories by using a limited yaw domain to circumvent angular singularities.
In this paper, we propose a novel \textit{global} yaw parameterization method for trajectory optimization that allows a 360-degree yaw variation as demanded by the underlying algorithm.
This approach effectively bypasses inherent singularities by including supplementary quadratic constraints and transforming the final decision variables into the desired state representation.
This method significantly reduces the needed control effort, and improves optimization feasibility.
Furthermore, we apply the method to several examples of different applications that require jointly optimizing over both the yaw and position trajectories. 
Ultimately, we present a comprehensive numerical analysis and evaluation of our proposed method in both simulation and real-world experiments. 

\end{abstract}

\IEEEpeerreviewmaketitle
\section{Introduction}

Unmanned Aerial Vehicles (UAVs) have a wide range of applications across various fields including industry, agriculture, search and rescue, and environmental monitoring\cite{7847361, lauri2016planning, 8633953, 8593416, 9325015}. 
These applications leverage the capabilities of UAVs to perform complex planning tasks. 
This is particularly challenging when considering the limited Field of View (FOV) of sensors on board such robots. 
The advancements in existing literature have extended trajectory generation to address attitude constraints for informative, obstacle-aware, and perception-aware navigation.

Conventional approaches \cite{liu2018towards, 9811688, 9531427, 9197157} typically address position and yaw commands independently, and provide control commands to ensure that the quadrotor consistently orients itself towards the target or along the current velocity vector.
Other works parameterize the one-dimensional yaw angles with smooth polynomials or splines~\cite{tao2023seer, zhou2021raptor}. 
However, it is challenging to get a globally consistent and uniquely defined attitude representation in spatio-temporal space for effective trajectory optimization.

The representation and parameterization of a robot's state play crucial roles in motion planning and control. 
While robot positions can be uniquely formulated in global coordinates, available representations for attitude can be problematic because of singularity and continuity issues.
The Euler angles vectors, quaternions, and rotation matrices are widely used in control within the local domain of definition but may introduce continuity or stability issues for global state space~\cite{5717652, hur2021effects}. 
High-dimensional rotation representations~\cite{Zhou2018OnTC} or methods~\cite{7039461} that use multiple charts to cover the $SO(3)$ manifold provide global parameterization; nevertheless, they significantly complicate the tasks of encoding perception-based constraints,  as well as minimizing the control effort in trajectory optimization.
The exponential map~\cite{doi:10.1080/10867651.1998.10487493} provides an almost global parameterization which can also be utilized for trajectory generation.
However, it needs an additional selection of the range of the angle bounds for the optimization initialization. 
The quality of the optimization is directly affected by such selections. 
In practice, finding a suitable selection for a particular application scenario is challenging. 

\begin{figure}[!t]
   \centering
   \vspace{0.2cm}
    \includegraphics[width=0.48\textwidth]{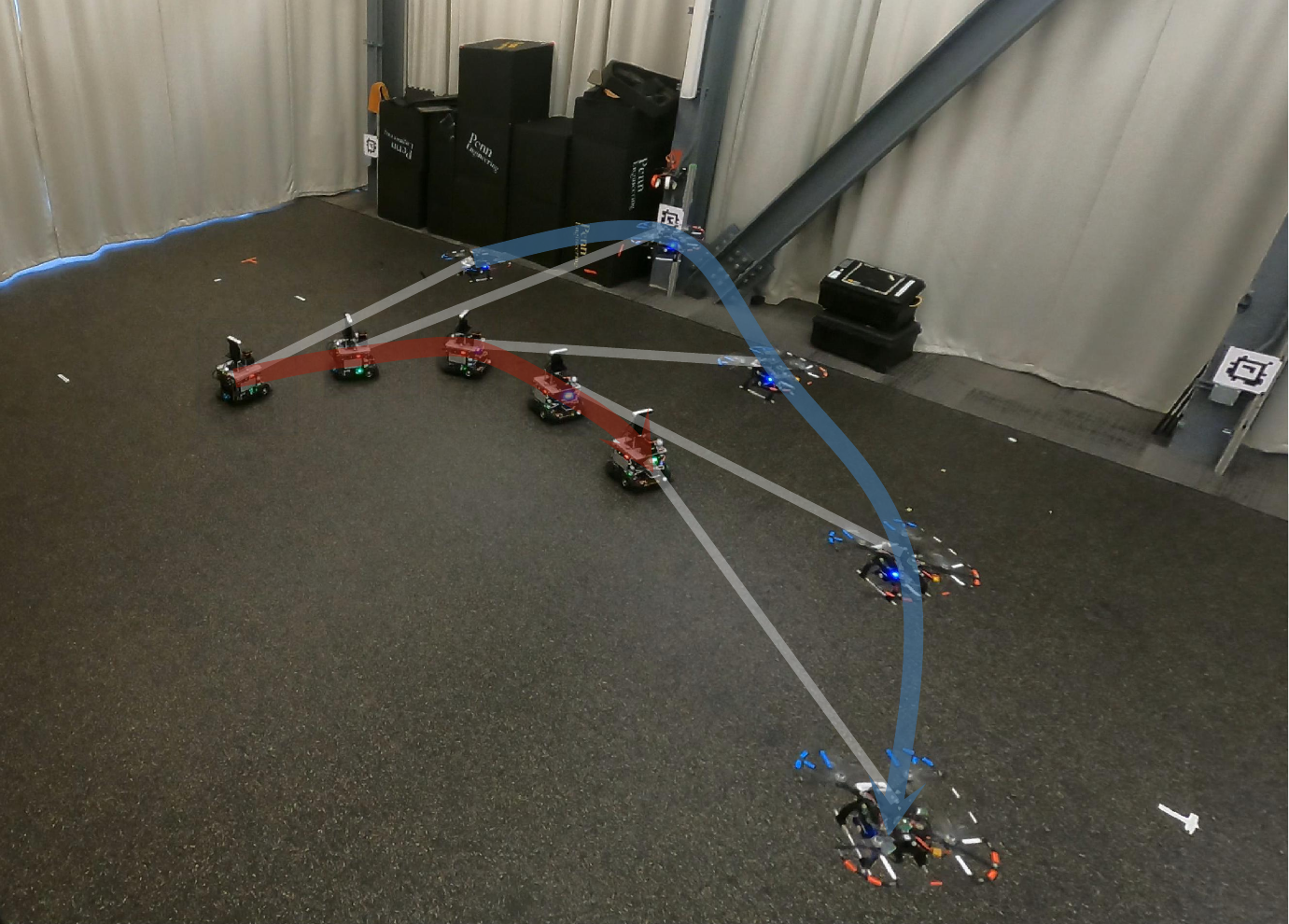}
   \caption{Snapshots of hardware experiment of traversal planning with yaw constraints. We use our SWaP-constrained Falcon250 quadrotor platform and Scarab ground robot platform as a tracking target (\ref{subsec:Hardware-Experiments}). The video is available at: \url{https://youtu.be/QMbkoTyKe-k} \label{fig: traversal}}
   \vspace{-0.7 cm}
   \end{figure}
The use of differential flatness with parameterized polynomials has addressed the challenges of representing nonlinear dynamics and generating trajectories in large-scale environments efficiently~\cite{5980409}. 
Flatness-based trajectory optimization can be initialized with intermediate keyframes and mapped to nonlinear dynamics to encode more complex constraints. 
However, the viewpoint constraints are often evaluated separately~\cite{zhou2021raptor, tao2023seer} given position trajectories.
Hence, a globally continuous trajectory optimization approach that couples robot positions and yaw angles is still a crucial challenge in real-world scenarios~\cite{8411468, tordesillas2022panther}. 

To bridge this gap, this work introduces a practical global yaw parameterization method for trajectory optimization. 
We represent yaw with suitable virtual variables represented as piece-wise polynomials, on which dynamic constraints are imposed.
Furthermore, we showcase the nonlinear problem formulation with constraints projected to real yaw angles. 
We unify trajectory generation problems for different applications and show that our proposed method can be efficiently incorporated into all of these scenarios. 
We present a thorough numerical analysis and experimental validation of our method, demonstrating its advantages and real-world applicability. Our contributions can be summarized as:

\begin{itemize}
   \item We propose a global yaw trajectory parameterization method for trajectory representation and optimization with orientation interpolation constraints.
   \item We develop a unified problem formulation and approach for different perception-based aerial tasks.
   \item We perform a thorough evaluation of the proposed method in simulation and carry out extensive hardware experiments of aerial tracking to validate the real-world applicability. 

\end{itemize}

\section{Related Work}

\subsection{Orientation Representations}

Effective full trajectory optimization relies on the representations and parameterization of all states.
While Euler angles are often convenient for representing orientation in control and planning, their use can introduce discontinuities that pose differentiation challenges during optimization.
Recent works\cite{1019463, 9197157, 9325015, DBLP:conf/rss/CohnPST23} have addressed this issue by optimizing on the local domain (of manifolds). 
These practical approaches involve constraining Euler angles within a 360-degree range to optimize limited horizon trajectories effectively.
To further ensure the dynamic feasibility of the trajectory, the direct incorporation of quaternions for path parameterization is proposed in \cite{Mao2023toppquad}. 
This approach introduces a distinct set of decision variables, allowing for the effective numerical approximation of quaternion dynamics along the specified geometric path.
More comparisons of various parameterization methods are provided in \cite{hur2021effects} on their application in direct feedback control.

However, the commonly applied quaternions and Euler angles are discontinuous in real Euclidean space and cannot achieve effective global parameterization. 
Utilizing alternative parameterization for rotations is a promising method to solve discontinuity, but finding concise representation in non-Euclidean spaces is challenging.
An exponential map is proposed in \cite{doi:10.1080/10867651.1998.10487493} for a more robust and continuous rotation representation. However, this approach still encounters singularities at intervals of  $2k\pi$. 
In \cite{1019463}, the authors proposed an SVD-based projection to generate trajectories for a rigid body and used the exponential coordinates for local parameterization.
In \cite{10.1023/A:1024265401576}, another rotation parameterization class utilizing geometric vectors in 3-D space was explored; still, it remains subject to a limited validity range.
Zhou et al.\cite{Zhou2018OnTC} demonstrate the continuity of rotations in higher dimensions (5-D or 6-D) to generate a continuous representation suitable for neural network training.
For trajectory parameterization and optimization formulation, the rotation representation is not efficient enough to encode complex perception constraints and is not able to handle infinitesimal variations\cite{doi:10.2514/1.G001041}.
\subsection{Yaw Trajectory Planning}

Leveraging the property of differential flatness, the complete dynamics of the quadrotor can be efficiently represented through flat outputs, and position and yaw trajectories can be parameterized by piecewise polynomials that can be optimized\cite{5980409}.
Perception-related constraints or objectives usually couple all flat outputs, introducing nonlinearity and modeling difficulties in the problem formulation.  
As position and yaw trajectories can be generated independently \add{with certain dynamic constraints} for tracking in the output space, separate yaw optimization with approximated perception constraints is extensively applied in many works.
Some previous works set the trajectory of yaw angles to values determined by a heuristic method, without optimizing it further~\cite{liu2018towards, 9811688, 9531427, tordesillas2022panther, 9197157, 9765821}.

By representing the yaw angle in real coordinate space, some previous works parameterize it with piece-wise polynomials \cite{zhou2021raptor,tao2023seer, 8411468, tordesillas2022panther} to formulate the minimum yaw control problem. 
In \cite{zhou2021raptor}, a perception-aware position trajectory is first generated with visibility constraints.  
Subsequently, a two-step planning method is adopted utilizing a graph search algorithm followed by local refinement to determine the optimal yaw sequence within $(-\pi, \pi]$. 
To further account for dynamics, yaw motion primitives are extended in \cite{tao2023seer} to generate a reference yaw sequence to avoid discontinuity of the solution set. 
The work in \cite{tordesillas2022panther} addressed the issues of decoupled optimization and proposed a coupled optimization using Hopf fibration. 
However, it still requires an initial guess of the sequence of yaw angles that is generated using a graph search method together with a shifting technique to ensure that consecutive elements of the sequence differ by less than $\pi$.
The discontinuity and singularities in yaw angles are usually fully addressed in these works. 
The singularities in differentially flat systems can be avoided through optimization on manifolds with constraints on coordinate charts \cite{doi:10.1177/0278364919891775} or utilizing a set of rotation charts\cite{7039461, kaminski2018intrinsic}.
However, these approaches are computationally expensive and lack the flexibility for motion planning scenarios with high dynamic range.
Our method can provide an efficient global yaw parameterization and planning algorithm based on differential flatness.

\section{keyframe Traversal Optimization}

\subsection{Notation}

We let $f^{(k)}$ denote $\frac{d^{k}}{dt^{k}}f$ for an arbitrary $k$-times differentiable function $f$, in addition to letting $\dot{f} \equiv f^{(1)}$. 
The unit circle $S^1 = \{ \mathbf{x} \in \mathbb{R}^2 \ \vert \ || \mathbf{x} ||^2 = 1 \}$ is represented as a subset of the (horizontal world) plane. 
Where necessary, ${}^{\mathcal{F}}p$ will denote coordinates of point $p$ with respect to the right-handed orthonormal reference frame $\mathcal{F}$.
The transformation between a pair of such frames $\mathcal{A}$ and $\mathcal{B}$ will be denoted by $({}^{\mathcal{A}}t_{\mathcal{B}}, {}^{\mathcal{A}}R_{\mathcal{B}})$ so that the relation ${}^{\mathcal{A}}p = {}^{\mathcal{A}}R_{\mathcal{B}} {}^{\mathcal{B}}p + {}^{\mathcal{A}}t_{\mathcal{B}}$ holds.

\subsection{Differential-flatness-based Trajectory Optimization}

\begin{definition}(Differential Flatness for Control-affine Systems\cite{fliess1995flatness}) A dynamical system of the form $\dot{x} = f(x) + g(x)u $ is said to have a flatness mapping if $x, u$ can be represented by the ``flat outputs'' and a finite number ($\kappa$) of their derivatives, $ (\mathbf{x}(t), \mathbf{u}(t) ) = \Psi(\bm{\sigma}^{[\kappa]}(t))$, where $ \bm{\sigma}^{[\kappa]}(t) = (\bm{\sigma}(t), \dot{\bm{\sigma}}(t), \cdots, \bm{\sigma}^{(\kappa)}(t) )$.
\end{definition}
We consider the flat outputs for a quadrotor system,
\begin{equation}
\bm{\sigma} = [p_x, p_y, p_z, \bm{\psi}]^T \in \mathbb{R}^3 \times S^1, 
\end{equation}
where $ p = [p_x, p_y, p_z]^T \in \mathbb{R}^3$ is the position of the quadrotor,  and $\bm{\psi} \in S^{1}$ is its heading vector. 
We define a smooth trajectory in flat space as $\bm{\sigma}: [t_0, t_M] \rightarrow \mathbb{R}^3 \times S^1$.
The minimum-control trajectory optimization problem can be formulated as follows
\begin{equation}
\begin{aligned}
\label{eq: opt1}
& \min_{\bm{\sigma}(\cdot)} J_c = \int_{t_0}^{t_M} \| \bm{\sigma}^{(\kappa)}(t) \|^2 dt, \  \quad  \quad  \quad  \quad \\
& \ {\rm s.t.} 		
 \ \mathcal{H}(\bm{\sigma}^{[\kappa -1]}(t) ) =  \bm{0}, \quad \forall t \in [t_0, t_M] , \\
& \ \qquad  \mathcal{G}(\bm{\sigma}^{[\kappa]}(t) ) \preceq \bm{0}, \quad   \forall t \in [t_0, t_M],
\end{aligned}
\end{equation}
where $\mathcal{H}$ defines equality constraints including boundary and continuity constraints between segments, and $\mathcal{G}$ defines customized inequality constraints. 
In previous works~\cite{5980409, tao2023seer}, the smooth trajectory of the heading vector is usually represented by a branch $\bm{\sigma}_{\psi} (t):  S^1 \mapsto \mathbb{R}$ that has discontinuities at integer multiples of $2\pi$; the latter cannot be defined globally via a differentiable function.

\subsection{Perception-based Trajectory Generation}

Autonomous navigation for a wide range of applications can be formulated as general trajectory optimization problems. 
In this context, we define \textit{keyframe} as a 3D position associated with a specific yaw angle\cite{5980409}. 
Subsequently, tasks such as inspection or exploration can be formulated as a \textit{keyframe} traversal planning problem as follows.

\begin{definition}(Keyframe Traversal Planning\cite{5980409}) Given a start state $x_{s}$, and a keyframe sequence $ \{ \bm{w_i} \in \mathbb{R}^n  | i = 1, \cdots, M \}$,  generate an optimal trajectory to visit these keyframes in order, with a criterion $\gamma(\cdot)$
\begin{equation}
\begin{aligned}
\bm{\sigma}^* = \argmin   \{ c(\bm{\sigma})  |  \ \bm{\sigma}(t_0) = x_{s} , \gamma (\bm{\sigma}(t_i) , \bm{w_i} ) \leq 0 ,  \\
i = 1, \cdots, M, t_i \in \mathbb{R}_{++}, \\
\bm{\sigma}: [t_0, t_M] \rightarrow \mathbb{R}^3 \times S^1 \},
\end{aligned}
\end{equation}
where $c(\cdot) $ is a cost function, and $ \gamma(\cdot)$ defines the function that evaluates if the keyframe is well-visited. 
\end{definition}
The objective of the exploration problem is to guide the robot through predefined keyframes while maintaining smooth commands and maximizing information gain. 
For aerial inspection, on the other hand, we have more flexibility to relax pose constraints, primarily focusing on ensuring that the static objects under inspection remain within the FOV of the robot. 
When dealing with constrained time intervals, the formulation can be expanded to encompass target-tracking problems defined below.

\begin{definition} (Time-constrained Traversal Planning)
Given a start state $x_{s}$, and a continuous function  $\{ \bm{w} : [t_0, t_M] \rightarrow  \mathbb{R}^3 \times S^1 \} $,  generate an optimal trajectory to visit these keyframes in order, with a criterion $\gamma(\cdot)$, 
\begin{equation}
\begin{aligned}
\bm{\sigma}^* = \argmin   \{ c(\bm{\sigma})  |  \ \bm{\sigma}(t_0) = x_{s} , \gamma (\bm{\sigma}(t) , \bm{w}(t)) \leq 0 , \\
\bm{\sigma}:[t_0, t_M] \ \mapsto \in \mathbb{R}^3 \times S^1 \}.
\end{aligned}
\end{equation}
\end{definition}
For target tracking and more complex scenarios like videography with dynamic obstacles, the trajectory is commonly analyzed through discrete evaluation, ensuring that the robot consistently maintains a distance from the target and has the target in its FOV\cite{9811688}.

\section{Methodology}

\subsection{Parameterization with Direct Mapping}

We represent the trajectory of heading vectors using the unnormalized (but nowhere-vanishing) virtual function $\bm{s}(\cdot)$ so that 
\begin{equation} \label{eq:virtual_to_heading}
\bm{\psi}(t) = \frac{\bm{s}(t)}{||\bm{s}(t)||}
\end{equation}
for all $t$.
The original flat outputs can then be encoded in a new space with redundancy, $[p_x, p_y, p_z, s_x, s_y]^T$. 
The conversion has one singular point at the origin, which can be avoided by ensuring that the distance $r = \| \bm{s} \| >0 $.

We can accordingly parameterize the new virtual trajectory with time as  $\bm{s}(t) = [s_x(t), s_y(t)]^T  \in \mathbb{R}^2 $, and map to the original heading vectors using the normalization function $g: \mathbb{R}^2 \setminus \{0\} \rightarrow S^1$ defined implicitly via
\begin{equation}
\bm{\psi}(t) = g(\bm{s}(t)) := \frac{\bm{s}(t)}{||\bm{s}(t)||}.
\end{equation}
Therefore, the minimal-yaw-control optimization problem can be formulated as:
\begin{equation}
\begin{aligned}
\label{eq: yaw_opt}
& \min_{\bm{s}(\cdot)}   \int_{t_0}^{t_M}   \|  g(\bm{s}(t)) ^{(\kappa)} \|^2 dt, \qquad \\
& \ {\rm s.t.}   \ \mathcal{H} \left (g(\bm{s}(t))^{[\kappa-1]} \right ) =  \bm{0}, \ \forall t \in [t_0, t_M], \\
& \ \qquad \mathcal{G}\left (g(\bm{s}(t))^{[\kappa]}\right )  \preceq \bm{0},  \ \forall t \in [t_0, t_M],
\end{aligned}
\end{equation}
where $\mathcal{G}(\cdot)$ represents dynamic and any perception-related constraints. The dynamic feasibility can be encoded as 
\begin{equation}
 \abs{g(\bm{s}(t)) ^{(1)}} \leq v_{\psi, \max}, \quad \abs{g(\bm{s}(t)) ^{(2)}} \leq a_{\psi, \max}, 
\end{equation}
where $v_{\psi, \max}, a_{\psi, \max}$ define the maximum angular velocity and acceleration. Any perception-aware, yaw value constraints can be encoded with a traversal criterion, 
\begin{equation}
\gamma (g(\bm{s}(t)), \bm{w}(t)) \leq 0 .
\end{equation}
Directly solving the initial problem (\ref{eq: yaw_opt}) would result in nonlinear constraints and objective functions. 
We will further discuss both convex optimization formulation via approximation for yaw trajectories and joint nonlinear optimization to efficiently generate coupled trajectories. 

\subsection{Optimization with Virtual Variables}

To generate a continuous and sufficiently smooth trajectory of yaw angles via quadratic programming, an alternative approach involves the direct optimization of the trajectory of virtual variables (see (\ref{eq:virtual_to_heading})).
To ensure smoothness, we paremeterize $\bm{s}(\cdot) $ as an $M-$segment piecewise polynomial function, with the coefficient matrix $\bm{c}_{\psi} = [ \bm{c}_{1, \psi}^T, \cdots, \bm{c}_{M, \psi}^T ]^T \in \mathbb{R}^{M (N+1)\times 2 }$, and time allocation intervals $\bm{t} = [ \Delta t_1, \cdots,  \Delta t_M]^T \in \mathbb{R}^{M}$, $\Delta t_i = t_{i} - t_{i-1}$, 
\begin{equation}
\begin{aligned}
\bm{s}(t) =   \bm{s}_i(t - & t_{i-1}) ,  \ \forall  t \in [t_{i-1} , t_{i}],  \\
\mathrm{where} \quad  \bm{s}_i(t) &= \bm{{\rm c}}_{i, \psi}^T \bm{\beta}(t),\\
     \bm{\beta}(t) &= [0, t, t^2, ..., t^N]^T, \  \forall  t \in [0, \Delta t_i].
\end{aligned}
\end{equation}
Given initial and terminal states, with a sequence of intermediate heading vectors $[ \bm{\psi}_1, ...,  \bm{\psi}_{M-1}]^T$, the minimum-yaw-control trajectory generation problem with dynamic bounds can be stated 
as 
\begin{subequations}
\begin{align}
\min_{\bm{s}(\cdot)} &  \int_{t_0}^{t_M}   \|  \bm{s}(t)^{(\kappa)} \|^2 dt   \\
{\rm s.t.} \ 
&\bm{s}^{[\kappa-1]}(0) = \Bar{\bm{s}}_0,\ \bm{s}^{[\kappa-1]}(t_M) = \bm{\Bar{\bm{s}}}_M,  \\
&\bm{s}(t_{i}) =  f(\bm{\psi}_i, r_i) ,\\
&\bm{s}_i^{[\kappa-1]}(\Delta t_i) = \bm{s}_{i+1}^{[\kappa-1]}(0) ,\\
& \| \bm{s}(t) ^{(1)}  \| \leq v_{s, \max}, \ \ \| \bm{s}(t) ^{(2)}  \| \leq a_{s,  \max}, \label{ineq: dynamic}\\
& \forall i \in {1, \cdots , M-1}, \ \ \forall  t \in [t_0, t_M] . 
 \end{align}
\end{subequations}
where $v_{s,  \max}, a_{s,  \max}$ represent dynamic bounds for virtual yaw variables, and $ \{r_1, \cdots, r_{M-1}\} \in \mathbb{R}_{++}$ is given for initialization. 
The function 
$f : S^1 \times \mathbb{R}_{++} \rightarrow \mathbb{R}^2, \ f: (\bm{\psi}, r) \mapsto r \bm{\psi}$
converts the heading vectors and radii into virtual variables.
The boundary states $\Bar{\bm{s}}_0, \Bar{\bm{s}}_M$ can be derived from initial and terminal heading vectors, angular velocities and radii $\{ \bm{\psi}_0, \dot{\bm{\psi}}_0,  r_0 \}, \{ \bm{\psi}_M, \dot{\bm{\psi}}_M,  r_M \}$.
The continuous-time constraints Eq. (\ref{ineq: dynamic}) are enforced at a set of discretization points along the trajectory. 

The singularity of the trajectory can be avoided via interactively inserting additional keyframes \cite{RicBryRoy1312} and checking the conditions of the radius.
We can further incorporate perception-related constraints between keyframes coupling the yaw and position trajectory optimization.

\subsection{Traversal Planning via Nonlinear Optimization}

More refined constraints would introduce non-convex constraints and non-convex terms in the objective function.
We will now discuss how the method described thus far can be extended to handle such refined constraints within a nonlinear optimization framework.
Considering the original joint optimization in Prob. (\ref{eq: opt1}) with only intermediate and boundary linear equations, the minimum control effort problem can be solved as a banded linear system with a unique solution. 
We use parameterized flat output with virtual variables $\bm{\sigma}_v(t) = [ \bm{p}(t), \bm{s}(t) ]^T \in \mathbb{R}^5$. 
Hence, we formulate an unconstrained nonlinear optimization problem in which perception constraints couple the trajectories of position and heading vectors of the robot in the form of
\begin{equation}
\min_{\bm{\sigma}_v(t)} J_c(\bm{\sigma}_v(t)) +  \sum_{i} \lambda_i I(\mathcal{G}_i(\bm{\sigma}_v(t), \cdot)),
\end{equation}
\add{where $J_c(\cdot)$ is the total control effort, and $\mathcal{G}_i$ represent different inequality constraints, each weighted by $\lambda_i$}.
The relaxation of inequality constraints can be achieved through the method presented in \cite{9765821}, applying a time integral penalty functional denoted as $I(\cdot)$.
We follow similar cost formulations for position-only constraints and focus on the discussion of perception-related penalties in the following context. 
For cost evaluation, we analyze the original yaw dynamics, 
\begin{equation}
    \dot{\bm{\psi}}(t) = 
    \left( {\rm I}_{2} - \frac{\bm{s}(t)\bm{s}(t)^T}{|| \bm{s}(t) ||^2} \right) \frac{\dot{\bm{s}}(t)}{|| \bm{s}(t) ||},
\end{equation}
and propagate the gradients of the cost functional with respect to the virtual variables $ (\bm{s}(t), \dot{\bm{s}}(t), \cdots, \bm{s}^{(\kappa-1)}(t) )$ using the chain rule.

\subsubsection{FOV Cost Functional}

The first condition evaluates the related position and orientation toward tracking objects, points of interest, and features within FOV to ensure it is in the image. 
The field of view region can be encoded with different models like a rectangular pyramid composed of an intersection of five half spaces \cite{9120168, 8814697}, or some partial spaces of a sphere \cite{9913848}. 
Hence, the penalty towards a point $\bm{w}(t)$ is defined as maximizing the total visibility,
\begin{equation}
\mathcal{G}_v (\bm{\sigma}_v(t), \bm{w}(t)) = - \mathcal{V}({}^{\mathcal{C}}\bm{w}(t)),
\end{equation}
where ${}^{\mathcal{C}}\bm{w}(t) = \bm{p}(t) + \mathbf{R}^T(t) \bm{w}(t)$ is the feature point in body frame, and $\mathcal{V}(\cdot)$ is a visibility function. 
An additional constant transformation matrix needs to be added if the camera frame is different from the body frame.
This formulation will introduce a rotation matrix during the optimization phase, which can be represented by flat output. 
To eliminate rotation, one can also penalize the relative distance of the point to the center of the image. 
The distortion and potential disturbance can also be penalized for vision-based sensors~\cite{falanga2018pampc}.
To guarantee that the object remains within the sensor's horizontal FOV ($\theta$), a relaxed cost can be employed
\begin{equation}
\mathcal{G}_v (\bm{\sigma}_v(t), \bm{w}(t)) = \cos \left(\frac{\theta}{2}  \right) \| \bm{p_r}(t) \| - \langle [\bm{\psi}(t)^T, 0]^T, \bm{p_r}(t)\rangle,
\end{equation}
where $\bm{p_r}(t) = \bm{p}(t) - \bm{w}(t) $ is the relative position to the center point of the object or features that the robot keeps tracking. 
When the tracking target is not a single point but multiple features, this term will become a multi-objective gain. As a result, the robot may not be able to keep all feature points in the field of view. 
For an environment with obstacles, rays between each tracking point to the robot should also be collision-free to avoid occlusion.
One way to guarantee safety is to ensure that the quadrotor should see the region before it reaches the region.
It can be practically formulated to enforce yaw angles toward the direction of the quadrotor's velocity  within a deviation threshold angle $\theta_r$, as
\begin{equation}
\mathcal{G}_v (\bm{\sigma}_v(t), \bm{w}(t)) = \cos(\theta_r) \| \dot{\bm{p}}(t) \|- \langle \bm{\psi}(t), \dot{\bm{p}}(t) \rangle. \\
\end{equation}
All of the above costs involve both position and yaw trajectories that can be jointly refined and optimized. 

\subsubsection{Velocity Cost Functional} 
To further ensure the perception quality, limiting related velocities towards features or tracking objects can also improve total perception quality\cite{falanga2018pampc}. 
The penalty on relative velocity of the point of interest ${}^{\mathcal{C}}\bm{w}(t)$ in the image plane coordinate is represented as
\begin{equation}
\mathcal{G}_r (\bm{\sigma}_v(t), \bm{w}(t)) = \left\| {}^{\mathcal{C}}\dot{\bm{w}}(t) \right\|^2 - v_{w,  \max}^2 , \\
\end{equation}
where $v_{w, \max}$ are the maximum velocity bounds for the feature point respectively.

\subsubsection{Dynamics Cost Functional}

By analyzing the original yaw dynamics, we can also evaluate the control efforts of the yaw trajectory.
The violation is defined as
\begin{equation}
\begin{aligned}
\left\{ 
\begin{array}{l}
\mathcal{G}_d  (\bm{\sigma}_v^{(1)}(t))  = \| \bm{\psi}^{(1)}(t) \|^2 - v^2_{\psi, \max} , \\
\mathcal{G}_d (\bm{\sigma}_v^{(2)}(t)) = \| \bm{\psi}^{(2)}(t) \|^2 - a^2_{\psi, \max}. \qquad \qquad  \\
\end{array}
\right. 
\end{aligned}
\end{equation}
In addition to adding cost to the original yaw angles, we also enforce the bounds of virtual variables to ensure the feasibility and continuity of the original yaw trajectory. 
We add radius constraints to avoid singularity,
\begin{equation}
\left\{ 
\begin{array}{l}
\mathcal{G}_d (\bm{\sigma}_v(t)) = r_{\min}^2 - \| \bm{s}(t) \|^2, \\
\mathcal{G}_d ( \bm{\sigma}_v^{(1)}(t)) = \| \bm{s}^{(1)}(t) \|^2 - v^2_{s, \max}, \\
\mathcal{G}_d ( \bm{\sigma}_v^{(2)}(t)) =  \| \bm{s}^{(2)}(t) \|^2 - a^2_{s, \max},
\end{array}
\right. 
\end{equation}
where $r_{\min} $ is the lowerbound for radius. 
The dynamic constraints can be evaluated with both the original yaw dynamics and the virtual variables.

\section{Results}

\subsection{Implementation Details}

We utilized 3rd-order polynomials to parameterize the piece-wise trajectories, while the evaluation of yaw angle control efforts is based on the integral of intensity of angular accelerations.
We use OSQP~\cite{osqp} to solve the problems for benchmark comparison. 
To address traversal planning with time-related perception constraints, we formulated the problem as a joint nonlinear optimization involving both position and yaw trajectories. 
We calculated the gradients of constraints and solved the nonlinear optimization by L-BFGS~\cite{liu1989limited}, using a framework from~\cite{9765821} with the proposed parameterization\footnote{The open source code will be available at: \url{https://github.com/KumarRobotics/kr_param_yaw}}.

\subsection{Benchmark Comparison for Traversal Planning}

We compared the proposed method with two other traditional parameterization methods for yaw angles.
The first method directly optimizes yaw angles in the range of $(-\pi, \pi]$ \cite{zhou2021raptor}. 
The second method uses yaw motion primitives \cite{tao2023seer} to incrementally construct a sequence of yaw values in $\mathbb{R}$, in which the newest element is chosen between the pair of angles closest to the last element of the current sequence that satisfies required heading interpolation constraints. 

\begin{figure}[!htp]
\centering
\vspace{0.1cm}
\includegraphics[width=0.48\textwidth]{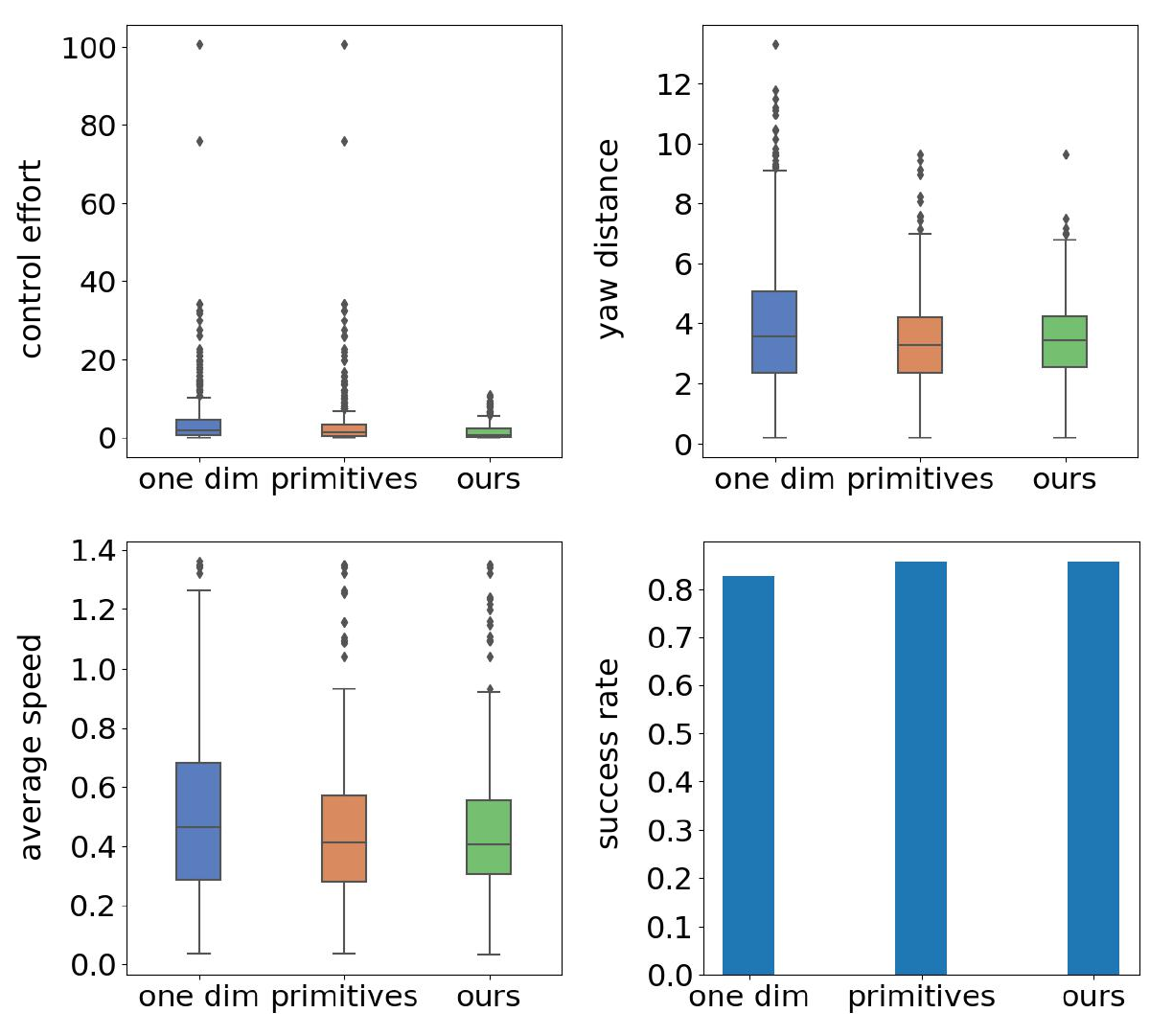}
\vspace{-0.3cm}
\caption{The comparison of different parameterization methods. We evaluated the relative minimum control cost ($ {\rm rad^2/s^3 }$), accumulated yaw distance (rad), average speed (rad/s), and success rate. \label{fig: benchmark}}
\vspace{-0.6cm}
\end{figure}

Given cluttered environments and evaluation pipeline from \cite{10610207}, we randomly generated 500 problem instances in $20 \times 20 \times 5 \  {\rm m^3 } $ random maps. 
Then we solved those problem instances using our method, as well as using the other two baseline alternatives for yaw planning given position trajectories.
We used the same time allocation in all methods to ensure a fair comparison.

As shown in Fig.\ref{fig: benchmark}, the first baseline method incurs a higher cost in terms of the control effort, accumulated yaw variations, and average speed.
The second baseline is able to avoid discontinuity issues, but it still requires a higher degree of control effort than the proposed method.
By giving feasible time allocations, all of these methods are able to achieve a relatively high success rate, while in some cases the first baseline fails to find a feasible solution as the moving direction of yaw angles takes more effort. 

\begin{figure}[!ht]
\vspace{0.2cm}
\centering
\includegraphics[width=0.48\textwidth]{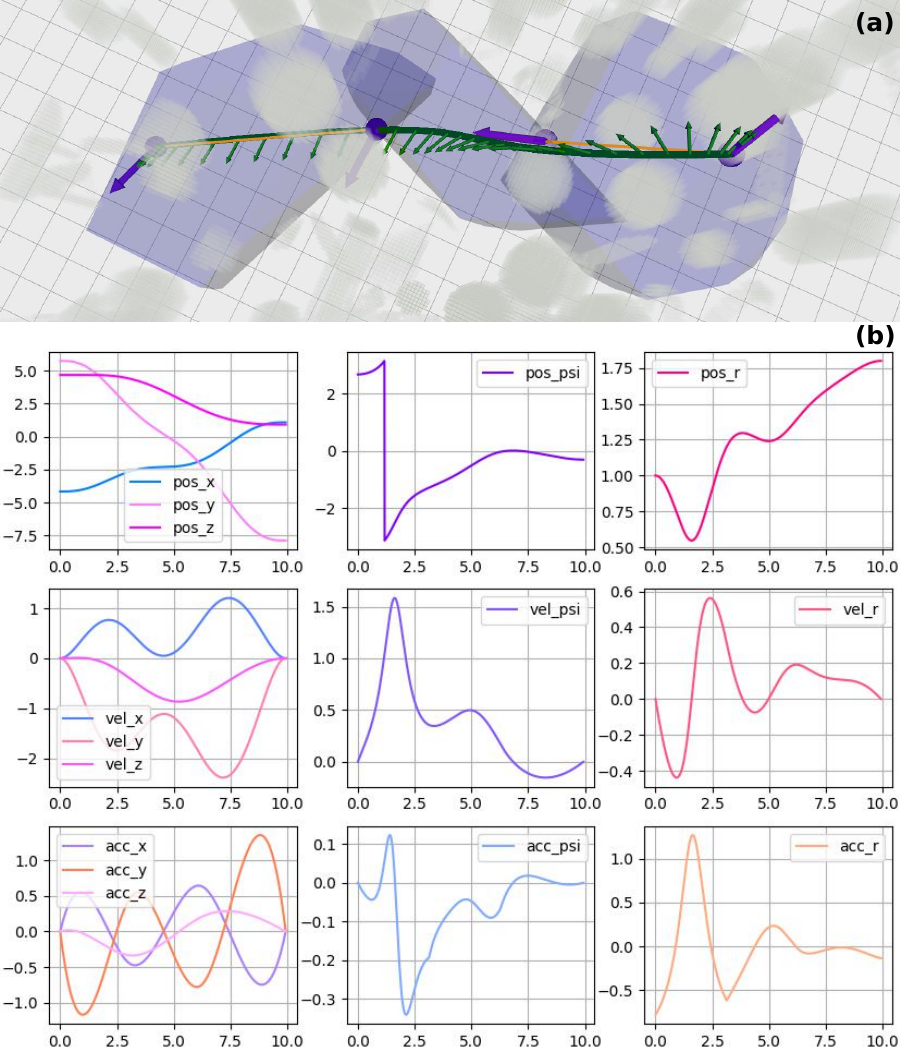}
\caption{ (a) Trajectory of proposed parameterization trajectories. The purple arrows represent position and yaw sequence, and the green arrows represent a sequence of discretized optimal trajectory.  (b) The plots of position, velocity, and accelerations of position, yaw, and r trajectories. \label{fig: yaw_traj}}
\vspace{-0.5cm}
\end{figure} 

In Fig. \ref{fig: yaw_traj}, we present a task with multiple intermediate interpolation constraints as an illustration.
By integrating with virtual variables, we can generate a smooth trajectory to avoid discontinuity and singularity.   

\subsection{A Note on an Alternative Representation }

Another global optimization can be formulated by applying nonlinear equality constraints to yaw angle $\psi \in \mathbb{R} $, as
\begin{subequations}
\begin{align}
\min_{\psi(t)} &  \int_{t_0}^{t_M}   \| \psi(t)^{(\kappa)} \|^2 dt,   \\
{\rm s.t.} \ 
&\psi^{[\kappa-1]}(0) = [\psi_0^{(0)}, ..., \psi_0^{(\kappa-1)}], \\
&[\psi^{(1)}(t_M), ..., \psi^{(\kappa-1)}(t_M) ]= [\psi_M^{(1)}, ..., \psi_M^{(\kappa-1)}],
\\
& \begin{pmatrix}
\cos(\psi(t_i))\\ 
\sin(\psi(t_i))
\end{pmatrix} \cdot \begin{pmatrix}
\cos(\psi_i))\\ 
\sin(\psi_i))
\end{pmatrix} 
= 1, \forall i \in {1, ... , M}, 
\\
&\psi_i^{[\kappa-1]}(\Delta t_i) = \psi_{i+1}^{[\kappa-1]}(0) , \ \forall i \in {1, ... , M-1}, \\
& \abs{\psi ^{(1)}(t) } \leq v_{\psi, \max}, \ \forall  t \in [t_0,  t_M], \\
&\abs{\psi ^{(2)}(t)}\leq a_{\psi, \max}, \
\forall  t \in [t_0,  t_M] . 
\end{align}
\end{subequations}
However, satisfying the nonlinear equality constraints is more challenging than the direct linear inverse, which makes it intractable for nonlinear solvers to generate reasonable results in the allowed time budget.

\subsection{Hardware Experiments}
\label{subsec:Hardware-Experiments}

For the application of traversal planning like exploration and inspection, extra constraints can be added accordingly. 
In real-world deployments, we focus on aerial tracking tasks to demonstrate time-constrained traversal planning.

We performed several real-world experiments on tracking moving targets to validate the efficacy of our proposed method in time-constrained tasks.
The quadrotor Falcon250~\cite{tao2023seer} has an Intel NUC 10 onboard computer with an i7-107100 CPU. 
The Scarab ground robot~\cite{Scarab2008} is used to serve as the moving target.
It has an onboard computer with an Intel i7-8700K CPU with 32 GB RAM, equipped with an Intel RealSense D455 RGB-D camera and a Hokuyo UTM-30LX laser. 
In all of our experiments, we ran GMapping and Human-Friendly Navigation (hfn)~\cite{hfn} on board the Scarab ground robot for SLAM and navigation. 
The Vicon Motion Capture System was used to set up the common reference frame. 
The quadrotor fetched its odometry and the ground robot's odometry from the Vicon for the traversal planning. 
The maximum velocities of the ground robot and the quadrotor were set to 0.5 $ {\rm m/s } $ and 1.0 $ {\rm m/s } $, respectively.
The desired distance between the robot and the target was set to 2.0 ($\pm$ 0.3) $ {\rm m } $, and the planner was running at 10 HZ onboard.

\begin{figure}[!ht]
\vspace{0.2cm}
\centering
\includegraphics[width=0.48\textwidth]{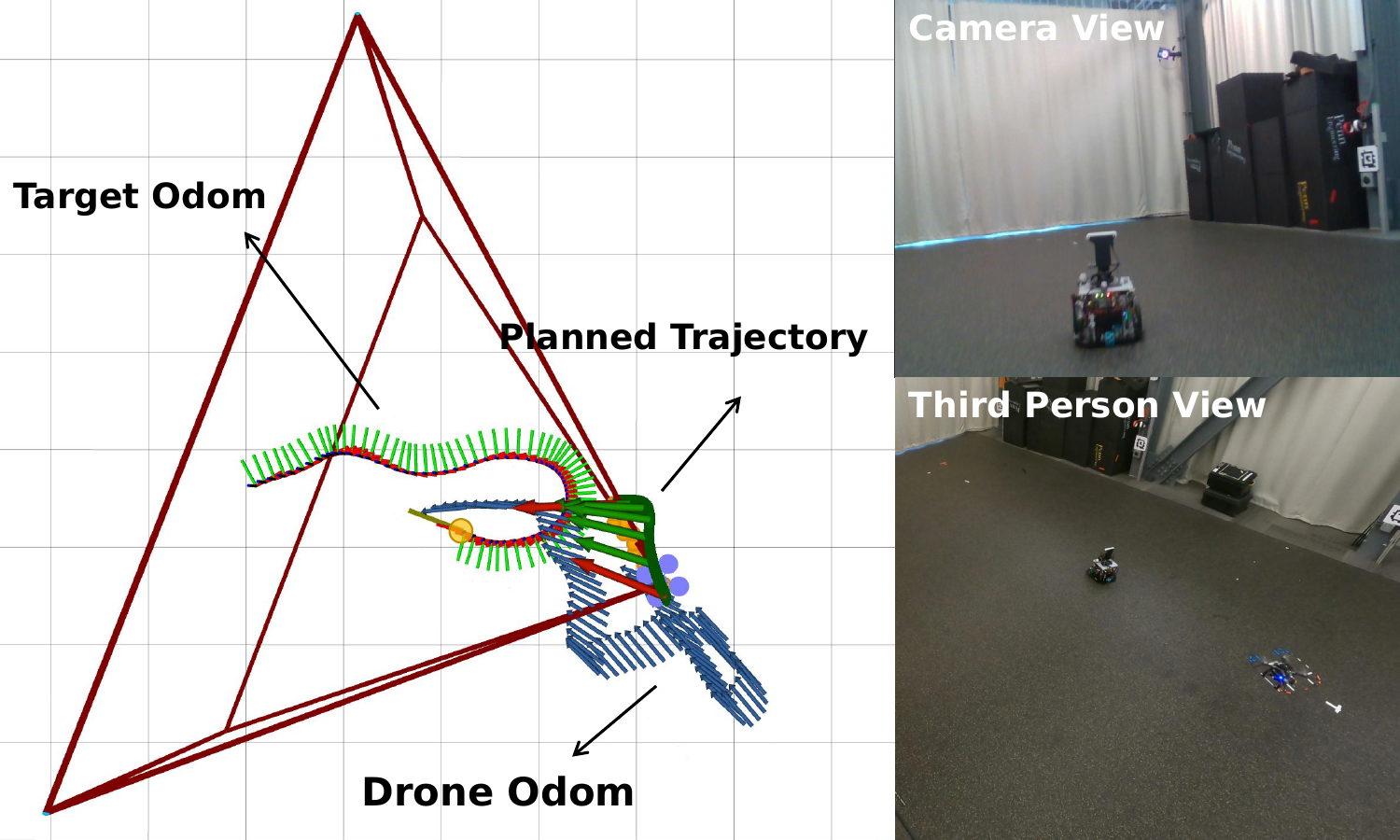}
\caption{Target tracking experiment. The right top image illustrates a current camera view from the quadrotor and the right bottom one shows a third person view. The left image visualizes its online planning process. The blue arrows represent the executed trajectory of the drone and the axes represent the odometry of the target. The red pyramid shape represents the current field of view. The red arrows represent the current start and end yaw angle, and the intermediate green arrows represent the planned trajectory. \label{fig: experiment}}
\vspace{-0.2cm}
\end{figure} 

The experiments are shown in Fig.~\ref{fig: traversal} and Fig.~\ref{fig: experiment}. 
As the ground robot moved around, the quadrotor captured a simulated observation of the ground robot from the motion capture system and predicted its future trajectories.
Subsequently, the quadrotor used the predicted target trajectories to generate visible keyframes and optimize the traversal trajectories which actively keeps the target within its FOV.  

\begin{table}[htbp]
    \renewcommand\arraystretch{1.4}
    \setlength{\tabcolsep}{6pt}
    \centering
     \vspace{-0.1cm}
    \caption{Evaluation of aerial tracking experiment \label{tab: exp}}
    \begin{tabular}{|c|cc|cc|cc|}
    \hline 
   Out of  FOV (\%)  &
    \multicolumn{2}{c|}{\tabincell{c}{Deviation \\ Angle (rad)} }   &  
   \multicolumn{2}{c|}{\tabincell{c}{  (z-Axis) Body   \\ Rate (rad/s)}}  &
   \multicolumn{2}{c|}{\tabincell{c}{Relative  \\ Distance (m)}} \\
    \hline 
   \multirow{2}{*}{2.07} & Avg & Std & Avg & Std & Avg & Std \\
   \cline{2-7} 
   &  0.29 &0.16 &  0.12 & 0.12 & 2.30 &0.36 \\
    \hline 
    \end{tabular}
    \vspace{-0.1cm}
\end{table}

The results of multiple real-world experiments are presented in Tab. \ref{tab: exp}. 
The quadrotor was able to keep the tracking target inside its FOV for more than 95\% of the time. 
The average relative angle between the target and the center of the robot's FOV was 0.29 (rad).
These statistics demonstrate the effectiveness of the proposed method in the aerial target-tracking task.

\section{Conclusion}\label{sec:conclusion}

In this work, we proposed a yaw parameterization method using flat outputs to jointly optimize the yaw and position trajectories. 
This approach relaxed yaw values in $S^1$ and practically avoided the singularity by adding extra quadratic constraints into the optimization problem.
We formulated the perception-based navigation application with uniform traversal planning and demonstrated the effectiveness of our proposed methods in a nonlinear joint optimization framework. 
We present comprehensive numerical analysis and experiments to validate our proposed method's advantages.
We envision our approach can be applied to various real-world applications.

\bibliography{references}
\end{document}